\documentclass[5p,twocolumn]{elsarticle_local}
\usepackage{latexsym}
\usepackage{graphicx}
\usepackage{amsfonts,amssymb,amsmath}
\usepackage{hyperref}

\def\BibTeX{{\rm B\kern-.05em{\sc i\kern-.025em b}\kern-.08em
    T\kern-.1667em\lower.7ex\hbox{E}\kern-.125emX}}

\DeclareUnicodeCharacter{0301}{\'{e}}

\usepackage{enumitem}

\usepackage{microtype}

\usepackage{changepage}

\usepackage{graphicx}

\usepackage{caption}
\usepackage{subcaption}
\usepackage{soul}

\usepackage[linesnumbered,ruled]{algorithm2e}
\usepackage{algorithmicx,algpseudocode}
\usepackage{ifthen}
\usepackage{amsfonts}

\usepackage{hyperref,xcolor}
\usepackage{color, colortbl}
\definecolor{Gray}{gray}{0.9}
\definecolor{Gray2}{gray}{0.85}
\definecolor{Gray3}{rgb}{0.94,0.94,0.94}
\definecolor{Blue}{rgb}{0.8,0.83,0.95}
\definecolor{LightBlue}{rgb}{0.929,0.964,0.992}
\definecolor{LightGreen}{rgb}{0.898,0.996,0.803} 

\usepackage{amsmath,amssymb}

\usepackage{dsfont}

\usepackage{booktabs}
\newlength{\Oldarrayrulewidth}

\usepackage{booktabs}
\usepackage{multirow}
\usepackage{dsfont}

\usepackage{comment}
\definecolor{Gray}{gray}{0.70}
\definecolor{Gray2}{gray}{0.90}
\definecolor{LightCyan}{rgb}{0.88,1,1}

\newcolumntype{b}{>{\columncolor{Gray}}c}
\newcolumntype{a}{>{\columncolor{Gray2}}c}
\newcolumntype{d}{>{\columncolor{LightCyan}}c}

\usepackage{lineno}
\begin{document}

\begin{frontmatter}

\title{Judge Me in Context: A Telematics-Based Driving Risk Prediction Framework in Presence of Weak Risk Labels}

%% Group authors per affiliation:
\author[osu]{Sobhan Moosavi}
\ead{moosavi.3@osu.edu}

\author[osu]{Rajiv Ramnath}
\ead{ramnath.6@osu.edu}

\address[osu]{Department of Computer Science and Engineering, The Ohio State University, Columbus, Ohio}

\begin{abstract}
Driving risk prediction has been a topic of much research over the past few decades to minimize driving risk and increase safety. 
The use of demographic information in risk prediction is a traditional solution with applications in insurance planning, however, it is difficult to capture true driving behavior via such coarse-grained factors. Therefor, the use of \textit{telematics data} has gained a widespread popularity over the past decade. While most of the existing studies leverage demographic information in addition to telematics data, our objective is to \textit{maximize} the use of telematics as well as \textit{contextual information} (e.g., road-type) to build a risk prediction framework with real-world applications. We contextualize telematics data in a variety of forms, and then use it to develop a risk classifier, assuming that there are some weak risk labels available (e.g., past traffic citation records). Before building a risk classifier though, we employ a novel data-driven process to ``augment'' weak risk labels. Extensive analysis and results based on real-world data from multiple major cities in the United States demonstrate usefulness of the proposed framework. 
\end{abstract}

\begin{keyword}
Telematics Data, Contextual Information, Label Refinement, Contextualization, Risk Classification
\end{keyword}

\end{frontmatter}

\section{Introduction}
\label{sec:introduction}
A recent trend in the auto insurance industry is to use \textit{telematics data} (i.e., data collected by OBD-II devices or smartphone applications) together with \textit{demographic information} (e.g., age, gender, and marital status) to determine driving risk and insurance policy rates. The important benefit of using telematics data is to better demonstrate \textit{natural driving behavior}, which cannot be achieved otherwise if only demographic information is considered. However, understanding how to utilize telematics data to best effect remains a challenging problem. 
While one group of existing studies has employed telematics data by summarizing them as \textit{coarse-grained} representations \cite{verbelen2018unravelling,hu2018advancing,ayuso2019improving,guillen2019use}, another group of studies sought to use such information in a more \textit{fine-grained} form \cite{guo2013individual,klauer2014distracted,zheng2014driving,joubert2016combining,zhang2017safedrive,gao2018claims,he2018profiling,wang2018you,chen2019driving}, with neither set of studies having made the case for which way leads to a best practice. 

The actions of a driver will change depending on the circumstance in which she is driving. Therefore we need \textit{contextual information} in addition to the driver's trajectory in order to characterize the driver appropriately (if we do not understand the context a driver may be characterized as safe although she drives the same way on rainy days, and is actually a risky driver). Examples of contextual properties that will affect driving include location, time, traffic, and weather conditions. However, several existing studies have failed to incorporate proper contextual information when predicting future risk \cite{guo2013individual,klauer2014distracted,zheng2014driving,zhang2017safedrive,gao2018claims,Wang2018cnn,wang2018you,chen2019driving}. 

A significant challenge is to actually \textit{define} driving risk, before we can \textit{measure} it. Many previous studies have defined driving risk as the likelihood of involvement in an accident, a serious traffic offense (e.g., driving under the influence or DUI), or, in general, filing an \textit{at-fault} insurance claim \cite{peck1983statistical,gebers1992identification,rajalin1994connection,gebers2003using,guo2013individual,klauer2014distracted,zheng2014driving,joubert2016combining,zhang2017safedrive,he2018profiling,gao2018claims,wang2018you,chen2019driving}. Other recent studies defined the risk based on the driver taking dangerous actions (with the actions being specified by human experts) \cite{dingus2006100,guo2013individual,klauer2014distracted,zheng2014driving}. Still other studies have employed human experts to label drivers with appropriate risk scores given their history of driving as derived from telematics data \cite{wang2018you}. Rather than categorically defining risk in these terms, we define risk differently as a \textit{factor} of, but not completely explained by, an individual's previous driving record, including accidents and traffic violations. This information is readily available for a large group of drivers, and it can be used to assess driving behavior, but it can only be used as \textit{partial ground-truth labels}\footnote{Traffic citations or accidents records alone cannot accurately indicate whether someone is a risky driver or not, and therefore, we consider them to be partial or weak ground-truth labels.}. 

In this paper, we propose a new driving risk prediction framework that utilizes telematics data together with contextual information. This framework uses a finer-grained way of representing telematics data in comparison to previous studies (\cite{joubert2016combining,zhang2017safedrive,gao2018claims,gao2018driving,he2018profiling,hu2018advancing,Wang2018cnn,wang2018you}), and utilizes a variety of environmental factors to contextualize telematics data. 
Our risk prediction framework consists of three major parts described as follows. The first part derives fine-grained representations of telematics data in ``context''. We directly encode contextual information while creating these representations, and employ road type (e.g., residential, urban, and highway area), road shape (e.g., turn segments), and period of day or daylight (i.e., day or night) as contextual information. Our representation comprises trajectory information, that is, attributes such as speed, acceleration, angular-speed, and change-of-angle between two consecutive observations. The shape of a road is also a property we encode in our representation. To extract this shape, we use a previously proposed rule-based algorithm that achieves over 97\% accuracy in detecting smooth and sharp turns in a trajectory \cite{moosavinejaddaryakenari2020telematics}. One example of the use of this as context is a representation that encodes the ``distribution of speed over high-speed road segments during a particular time of the day''. 

The second part of our approach refines the above partial (or weak) risk labels. To do this, we compare each driver's  behavior against a set of ``presumably'' safe drivers (i.e., drivers with no past traffic records over a period of 10 years), and build a new representation which we call the \textit{deviation view}. We cluster deviation views to identify major risk cohorts. Next, we label each cohort based on the partial labels of their members with broader categories (low-risk, medium-risk, and high-risk). This cohort label is then transferred to each driver. 

Lastly, we train a risk cohort prediction classifier that uses deviation maps as input, and the drivers' cohort risk labels as the target. We experimented with several well-known classification models, ranging from traditional (e.g., Gradient Boosting Classifier) to deep models (e.g., Convolutional Neural Networks). Finally, we tested our approach on a large, real-world dataset of telematics data collected from five major cities in the United States (i.e., Philadelphia and Pittsburgh in Pennsylvania, Memphis in Tennessee, Atlanta in Georgia, and Columbus in Ohio)\footnote{This dataset is provided to us by our industrial sponsor, which is a major insurance company in Columbus Ohio.}. 

Our analysis and results point to significant benefits with contextualizing telematics data to improve risk prediction outcomes, and the use of our risk label refinement process to identify meaningful risk cohorts. Additionally, we observed that the use of a deep neural network model results in better risk prediction outcomes. We believe this lies in how we encode and represent the input data in terms of multi-dimensional matrices, which results in extracting more meaningful spatiotemporal patterns when a deep model (such as convolutional neural network) is employed. 

\iffalse
The main contributions of this paper are summarized as follows. 
\begin{itemize}
    \item A novel technique to contextualize telematics data by employing various sources of contextual information including road type, road shape, and daylight information. 
    %\item A highly-accurate rule-based turn detection algorithm that achieves over 97\% accuracy in identifying sharp and smooth turns when tested on real-world data. 
    \item A clustering-based technique to refine weak driving risk labels prior to building a risk prediction model. 
    \item An end-to-end driving risk prediction process to be employed for applications such as usage-based-insurance and driver coaching. 
    %\item A demonstration of the effectiveness of our approach through extensive experiments on a real-world dataset of telematics data with partial (or weak) risk labels.  
\end{itemize}
\fi

This paper makes the following main contributions:
\begin{itemize}
    \item A novel technique for contextualizing telematics data by incorporating various sources of contextual information, such as road type, road shape, and daylight information.
    \item A clustering-based approach for refining weak driving risk labels before building a risk prediction model.
    \item An end-to-end driving risk prediction process suitable for applications such as usage-based insurance and driver coaching.
\end{itemize}
\section{Research Problem}
\label{sec:problem}
Let us suppose that telematics data is composed of trajectories. A trajectory $T$ is a time-ordered sequence of data points $\langle p_1, p_2, \dots, p_{|T|}\rangle$. We represent each data point $p$ by a tuple $\big(t, Speed, Heading, Lat, Lng\big)$, where $t$ is a timestamp, $Speed$ is the ground velocity of a vehicle (in $km/h$), $Heading$ is the heading (or bearing) of a vehicle (number between 0 and 359 degrees, and $Lat$ and $Lng$ represent latitude and longitude. Further, any data that can be used in conjunctions with telematics data to better demonstrate environmental factors are called ``contextual information`` in this paper. Examples are traffic flow, daylight, and road characteristics (e.g., curvature). 
We formulate the risk prediction problem as follows: 

\vspace{5pt}
\noindent\textbf{Given:}
\vspace{-3pt}
\begin{itemize}
    \item [-] A set of drivers $\mathcal{D}$, with 
    \item [-] The \textit{observed risk score} for each driver $D \in \mathcal{D}$ denoted by $R_D$. Here, $R_D$ represents the total number of traffic accidents and violation records for driver $D$ in the past five years. %\textcolor{red}{RR: Is this risk score normalized in any way?}
    \item [-] A set of trajectories $\mathcal{T}_D$ for each driver $D \in \mathcal{D}$. 
    \item [-] Contextual information $C$ that could be used in conjunction with the trajectory data. 
\end{itemize}

\vspace{5pt}
\noindent\textbf{Build:}
\vspace{-3pt}
\begin{itemize}
    \item [-] A contextualized telematics representation $\mathcal{C}_D$ for driver $D \in \mathcal{D}$ based on $\mathcal{T}_D$ and accompanying context information $C$ (e.g., road shape, road type, and time information). 
    \item [-] A classifier $\mathds{C}$ to predict a risk cohort label $\mathcal{L}_D$ for driver $D \in \mathcal{D}$, using the representation above $\mathcal{C}_D$. Here, risk cohorts are categorical labels (such as \textit{low-risk} and \textit{high-risk}). 
\end{itemize}

\vspace{5pt}
\noindent\textbf{Objective:}
\vspace{-3pt}
\begin{itemize}
    \item [-] Minimize the average risk score for the low-risk cohort and maximize it for the high-risk cohort, while maintaining a size balance among cohorts. 
\end{itemize}

% \textcolor{red}{RR: Why a size balance? Is this reflective of reality?}

We note that the extension of this objective to more than two risk cohorts is straightforward. Also, the condition on balance between driver cohorts is to ensure to not end up with cases where a handful of safe (or risky) drivers belong in one cohort and the rest in the other.  
\section{Related Work}
\label{sec:related}
Driving risk prediction has been a topic of much research over the past few decades. While earlier studies mostly employed demographic information (e.g., driver's age, gender, marital status, car's age, and car's value)  \cite{peck1983statistical,gebers1992identification,rajalin1994connection,hu1998crash,lourens1999annual,elliott2000persistence,gebers2003using}, the recent studies have mostly used telematics data in addition to demographic information for driving risk prediction  \cite{guo2013individual,klauer2014distracted,zheng2014driving,moosavi2017characterizing,Wang2018cnn,gao2018claims,wang2018you,ayuso2019improving,hu2018advancing,guillen2019use,pesantez2019predicting,denuit2019multivariate,perez2019quantile,eling2020impact,guillen2021percentile,maillart2021toward,moosavi2021driving,singh2021analyzing,so2021cost}. In this section. We provide an overview of related work by categorizing them based on the type of their input data. 

\subsection{Risk Prediction based on Demographic Data}
A large body of research on driving risk prediction has mainly relied on demographic data. 
Peck and Kuan \cite{peck1983statistical} proposed a multi-regression model that used data on accidents and violations from the past three years, as well as demographic factors, to predict the number of future accidents for a large group of drivers in California. They found that prior citation frequency, territorial accident rate, and prior accident frequency were the top three important factors in predicting future accidents. In another study, Gebers and Peck \cite{gebers1992identification} examined the risk of older drivers through several experiments. They found that traffic conviction record for older drivers is a stronger predictive factor for future accidents in comparison to younger drivers. Lourens et al. \cite{lourens1999annual} studied the impact of several factors on the risk of involvement in accidents. Using single and multivariate regression analyses, they found that annual mileage, previous traffic conviction records, age, age-citation\footnote{The driver's age and citation history taken together as a single factor.}, and gender-education to be significant predictor factors of risk. 
Separating experienced and non-experienced drivers in studying the impact of previous traffic convictions on future crashes or serious violations was examined by Elliot et al. \cite{elliott2000persistence}. They found that a serious offense in the previous year doubled the odds for a serious offense in the subsequent year, and a previous at-fault record increases the odds by 50\% for new at-fault crashes. 
A common aspect of all of these studies is that they have relied on demographic data, previous traffic conviction records, or questionnaire surveys to describe drivers, which may not accurately represent the naturalistic behavior of a driver. Without telematics data, fair and accurate prediction of driving risk is not possible due to inherent biases in applying predictions for a group of drivers to individual drivers and inadequate information about the past history of drivers. Furthermore, traditional approaches are limited in providing precise case-by-case prediction of driving risk. As a result, most studies have focused on analyzing and predicting risks for groups of drivers \cite{gebers1992identification,rajalin1994connection,hu1998crash,lourens1999annual,elliott2000persistence}. 

\subsection{Risk Prediction based on Telematics Data}
Using telematics data alone for driving risk prediction is a recent trend given notable attention in the past few years. 
Joubert et al. \cite{joubert2016combining} proposed using accelerometer and speed data as telematics for driving risk prediction. They used a two-dimensional matrix that only included accelerometer data along X and Y axes. The matrix was used to calculate a distribution of values for different cells, and risk categories were defined based on these values. The authors also incorporated speed and road type data to create matrices representing acceleration-based driving behavior for various types of roads and speeds. However, the study had limitations in terms of the amount and type of telematics and contextual data used. 
Hu et al. \cite{hu2018advancing} used telematics data for predicting future crash risk by analyzing patterns such as ``fast-acceleratio'', ``hard-braking'', and ``speeding'' in a specific context using regression-based models. They also performed map-matching to incorporate speed-limit and expected traffic speed data. However, they did not use finer-grained telematics data in their study. 
Wang et al. \cite{Wang2018cnn} proposed using a CNN model to predict driving risk from accelerometer data. They created a scatter map from the raw data and converted it into a 2D image. Then, they used a CNN-based auto-encoder to initialize their pipeline and train a second CNN network to predict risk as a binary label. 
In another study, Zhang et al. \cite{zhang2017safedrive} proposed a model to use telematics data to create state graphs, which can be used to detect abnormal driving behavior for individual drivers. They compared an individual's state graph with an aggregate state graph created from data of a large group of drivers to identify any unusual behavior. The authors also introduced pre-defined labels to classify the type of abnormal behavior, such as rapid acceleration, sudden braking, and rapid swerving. 
Wang et al. \cite{wang2018you} also used the state-graph representation to predict driving risk, but they combined it with an auto-encoder neural network model to derive a more compact representation. They used this compact representation to predict driving risk using a Support Vector Regression model. However, their method required manual labeling of drivers as risky or safe, which may not be a scalable solution. 

Guo and Fang \cite{guo2013individual} analyzed the 100-car naturalistic driving study (NDS) dataset to perform risk assessment for individual drivers based on crash and near-crash (CNC) and critical-incident events (CIE). Using negative binomial regression and K-Means clustering, they identified important risk factors such as age, personality, and CIE, and classified drivers into low, moderate, and high-risk clusters. Klauer et al. \cite{klauer2014distracted} used the same dataset to study the impact of secondary tasks on the risk of crash or near-crash events, and found involvement in tasks such as texting, eating, and using a cell phone increased the risk for both novice and experienced drivers. Chen et al. \cite{chen2019driving} also used the NDS dataset and developed an auto-encoder model to optimize input data window-size and a deep-neural-network to predict CNC events for near-crash and crash event prediction. 

\subsection{Risk Prediction based on Demographic and Telematics Data}
The last group of studies employed both telematics and demographic information for driving risk prediction. 
Gao et al. \cite{gao2018claims} used distribution matrices of speed and acceleration values to propose a framework that predicts the number of insurance claims by employing a Poisson General Additive Model (GAM). They condensed the telematics data representation and used driver and vehicle demographics as input parameters. In a follow-up study, Gao et al. \cite{gao2018driving} improved on their framework by extending demographic factors and representing telematics data with a wider range of speed values. They used a GAM model to predict the frequency of claims and concluded that lower speed ranges offered better predictivity for driving risk. 
Ayuso et al. \cite{ayuso2019improving} used telematics and demographic data to predict insurance claim frequency. Coarse-grained telematics factors included annual distance driven, distance driven at night, distance over the speed limit, and urban distance. They employed a Poisson regression model and found age, vehicle type, distance driven over the speed limit and urban distance to be the most significant factors. Guillen et al. \cite{guillen2019use} used a Zero Inflated Poisson regression model to correct for biases and avoid risk prediction based solely on the number of miles driven. Their coarse-grained telematics factors included distance driven over the speed limit, urban distance, and distance driven per year. Verbelen et al. \cite{verbelen2018unravelling} used regression-based models with coarse-grained telematics factors such as yearly distance, number of trips, and coverage of different road types. He et al. \cite{he2018profiling} proposed a comprehensive framework for risk prediction that extracted fine-grained information from trajectories using driving state variables and a machine learning model. However, a limitation of all these studies was the limited usage of contextual information. 

\subsection{Related Work Summary}
Our risk prediction framework falls into the category of using only telematics data to predict driving risk. However, our approach differs from existing solutions in several important ways. Firstly, we use finer-grained representations of telematics data to better capture driving behavior. Secondly, we incorporate contextual data to more accurately represent and evaluate driving behavior in context. Finally, we address the issue of weak risk labels by augmenting them before modeling and prediction. This is often overlooked when using traffic records, such as accident and claim data, as ground-truth labels to train or test risk prediction models. We recognize that traffic records or their absence alone are insufficient indicators of risk, and correcting this bias is necessary to build accurate risk prediction models.
\section{Methodology}
\label{sec:method}
\begin{figure}[t]
    \centering
    \includegraphics[scale=0.33]{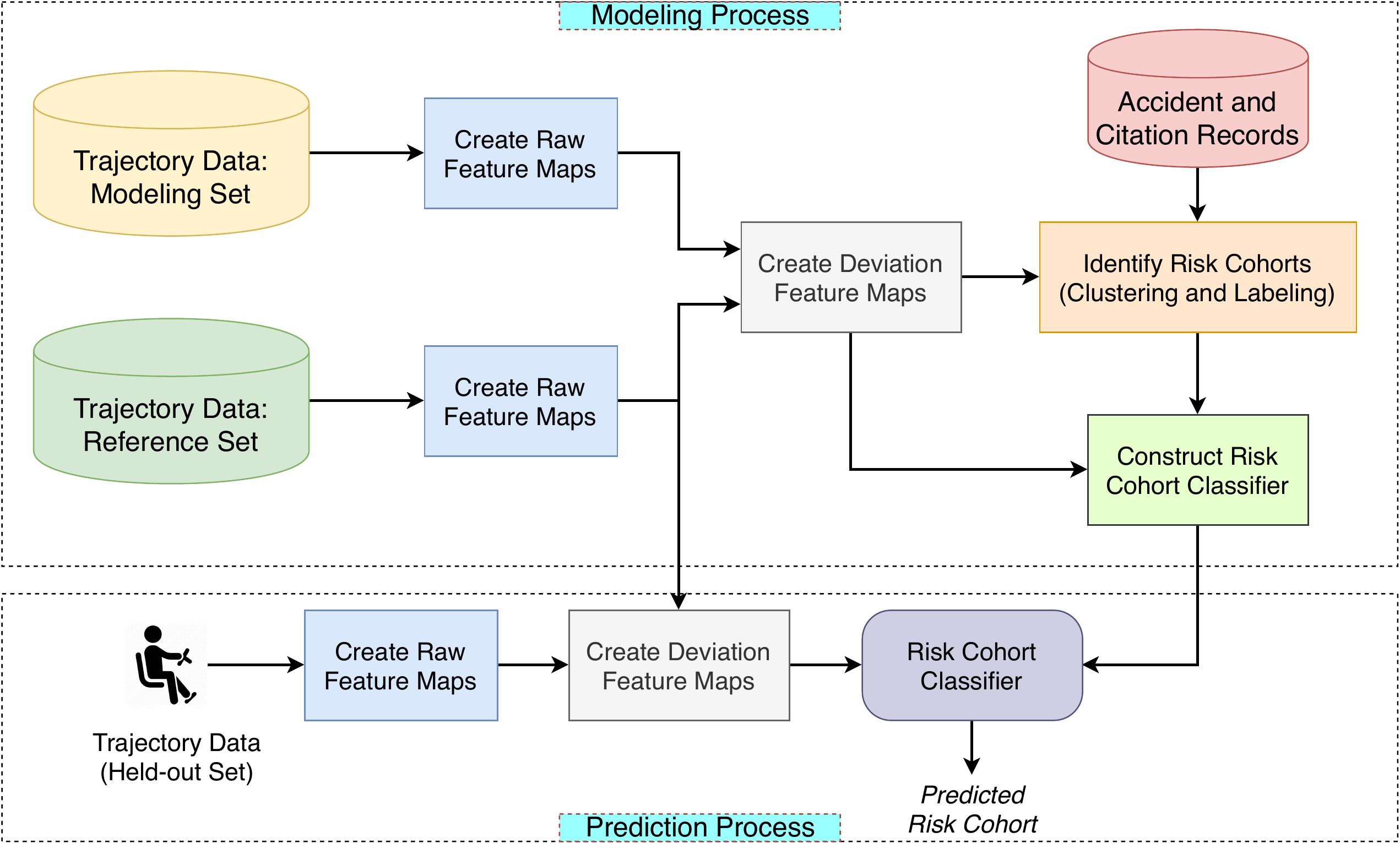}
    \caption{The proposed risk prediction framework. The ``Modeling Process'' builds contextualized telematics representation for drivers and constructs a risk-cohort classifier; and the ''Prediction Process'' predicts risk cohort for new or unseen drivers.}
    \label{fig:risk_prediction}
\end{figure}

This section describes our context-aware driving risk prediction framework that consists of three components. The first component creates a representation for telematics data in context. The second component builds a risk cohort classifier based on contextualized telematics information and weak (or partial) risk labels. Finally, the third component performs risk cohort prediction for unseen drivers in real-time as a real-world application. Figure~\ref{fig:risk_prediction} represents the big-picture of our approach, where the first two components make up the \textit{Modeling Process} and the third component makes up the \textit{Prediction Process}.

\subsection{Contextualized Telematics Representation}
In this section, we describe how to use telematics and contextual data to build representations for drivers. The telematics data consists of trajectory datasets and the representations are built based on different contextual conditions (e.g., road type and road shape). We next describe trajectory datasets, and then describe how to build representations for each trajectory. Lastly, we describe how to build representations for each driver. 

\subsubsection{Modeling Trajectory Dataset}
The modeling dataset consists of a large set of drivers (e.g., 4,000 drivers) with a minimum number of trajectories (e.g., 100 trajectories). These drivers must satisfy the following constraints: 1) each driver must own exactly one vehicle; and 2) the driver should not share her vehicle with any other driver. These constraints help ensure that trajectories collected for a vehicle had been generated by a single driver, who only drove that vehicle; thereby reducing the chance of vehicle-related biases\footnote{A driver might exhibit different behavior when driving different cars.}. We also refer to this set as the \textit{observation set} in this paper. 

\subsubsection{Reference Trajectory Dataset}
The reference trajectory dataset consists of a large set of \textit{presumably safe} drivers (e.g., 5,000 drivers), also with a minimum number of trajectories for each driver. Drivers in this set must have no record of accidents or traffic violations in the past 10 years, hence may be considered safe. %Note that having any conviction records is not necessarily an indicator of being risky. Additionally, not having such records does not necessarily endorse safeness. However, we heuristically assume that a driver with no records for a \textit{sufficiently} long period of time is a safe driver\footnote{Please note that drivers in this set were licensed for over 10 years.}. 

\begin{figure}[ht!]
    \small
    \centering
    \minipage{0.53\textwidth}
        \centering
        \includegraphics[width=\linewidth]{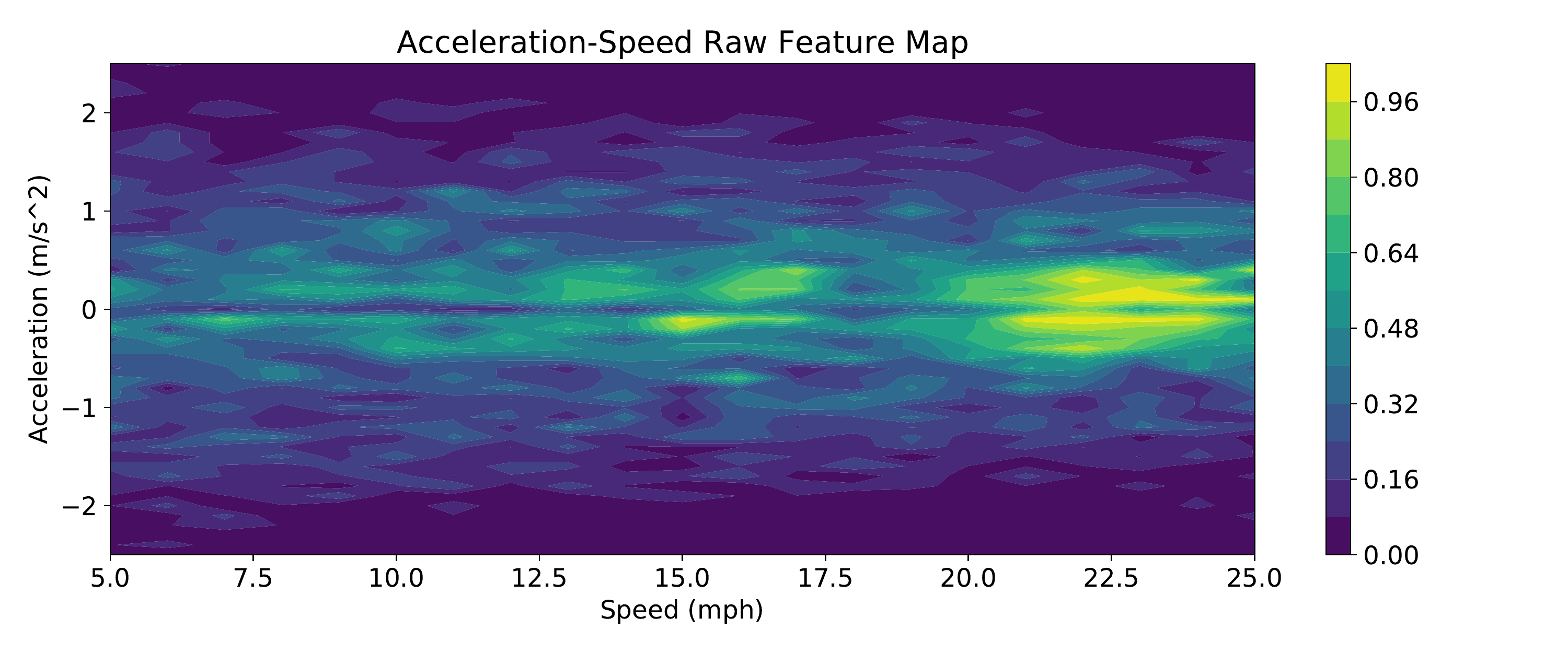}\vspace{-2pt}
        (a) Raw feature map (speed and acceleration)
        \label{fig:fm_spd_acl}
    \endminipage\\
    \minipage{0.53\textwidth}
        \centering
        \includegraphics[width=\linewidth]{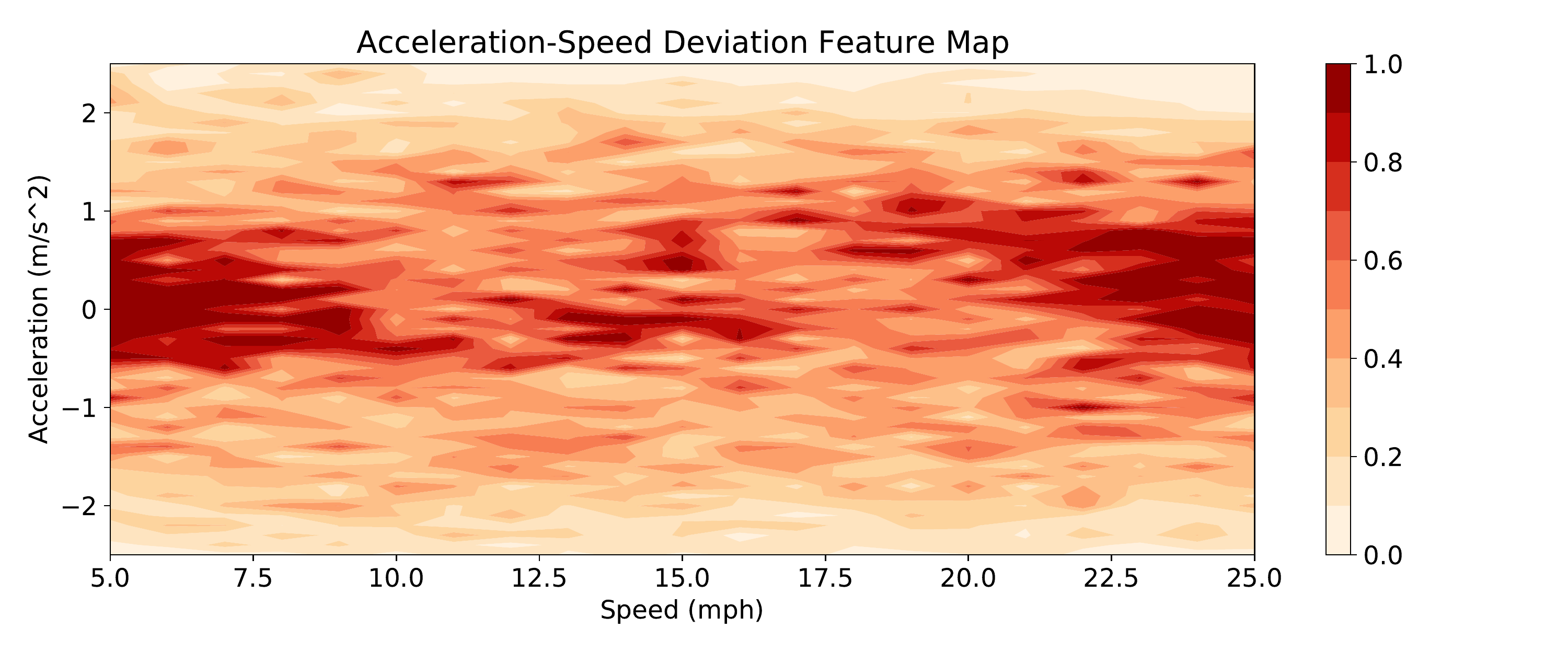}\vspace{-2pt}
        (b) Deviation feature map (speed and acceleration)
        \label{fig:fm_acl_anglchange}
        %\vspace{10pt}
    \endminipage
    \caption{Example of a ``raw'' and a ``deviation'' feature map for a trajectory and a driver, respectively}
    \label{fig:raw_dev_feature_maps}
\end{figure}

\subsubsection{Raw Feature Map: Representing a Trajectory}
\label{subsub:raw_feature_map}
The first step to using telematics data is to design a way to represent trajectories. We represent a trajectory using a \textit{feature map representation}. A feature map is a matrix, such that each of its axes shows a specific attribute (e.g., speed, acceleration, or angular speed), and each cell represents the \textit{normalized frequency} of trajectory data points that belong in that cell. As an example, suppose that we have a feature map where its x- and y-axis are speed and acceleration, respectively. A cell in this matrix represents specific speed and acceleration intervals (e.g., speed interval = [5, 25] $km/h$, and acceleration interval = [0.0, 0.2] $m/s^2$), and its value is the normalized frequency of trajectory data points whose speed and acceleration values fall in these intervals (where the raw frequency is normalized by the maximum frequency value). Figure~\ref{fig:raw_dev_feature_maps}-a shows an example of a raw feature map. %Figure~\ref{fig:raw_feature_maps}-a is a feature map for a single trajectory with speed and acceleration as the axes. Figure~\ref{fig:raw_feature_maps}-b shows another feature map generated for the same trajectory whose axes are acceleration and angle-change (or change of heading). Finally, Figure~\ref{fig:raw_feature_maps}-c shows another view of the same trajectory with axes as angular-speed and angle-change. Note that the last two raw feature maps are generated only for \textit{turn-segments} extracted from the represented trajectory, and are examples of utilizing contextual information to generate the representation for a trajectory. %More details on our turn detection process are provided in Section~\ref{sec:turn_detection}. 

\subsubsection{Deviation Feature Map: Representing a Driver}
\begin{figure}[h!]
    \centering
    \includegraphics[scale=0.7]{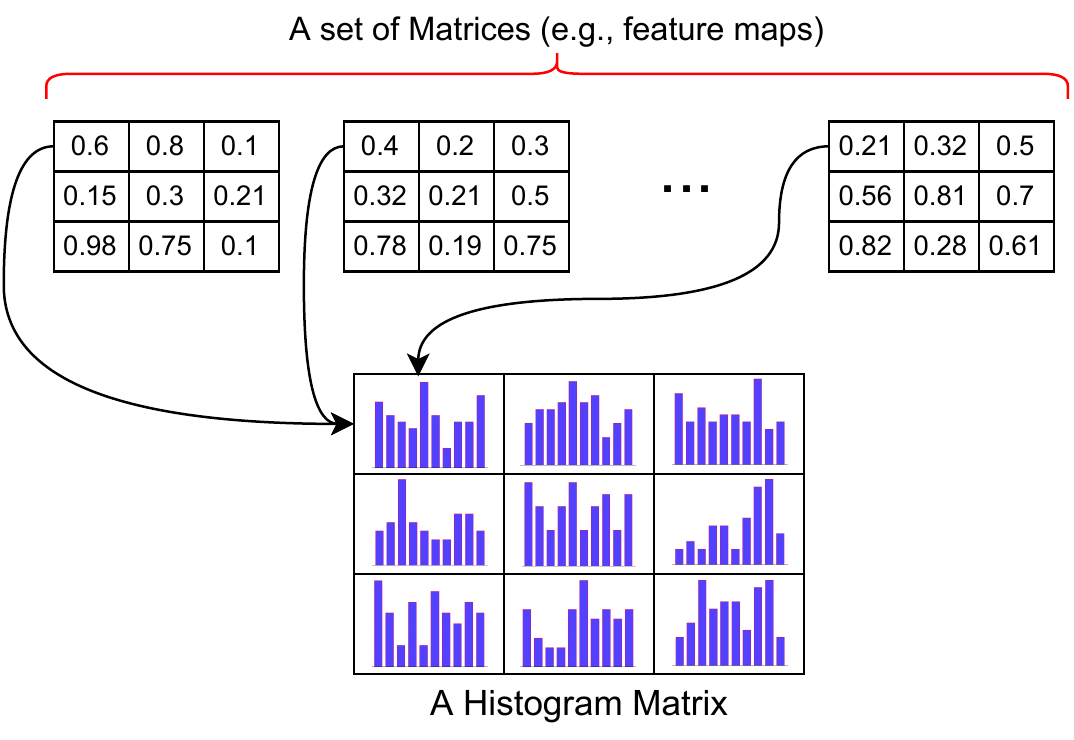}
    \caption{Example of a histogram matrix, created based on a set of input matrices. Each cell in histogram matrix represents data distribution across corresponding cells in input matrices.}
    \label{fig:histo_matrix}
\end{figure}

Given a set of raw feature maps generated for a set of trajectories taken from a driver $d$ (sampled from the observation set), we now describe our novel process to create a new representation for driver $d$, termed a \textit{deviation feature map}, and use it for predicting whether the driver is safe or risky. The main steps of this process are described by Algorithm~\ref{algo:dev_feature_map}. For this process, we randomly divide our reference set into \textit{base} and \textit{control} sets; such that the former set contains trajectories taken from $70\%$ of the presumably safe drivers, and the latter contains trajectories taken from the rest of the drivers in the reference set. % Note that we randomly sample drivers from the Reference set (without replacement) and create these sets. 
The process starts by generating raw feature maps for trajectories in the base set (lines 1--7). Then, we generate a \textit{histogram matrix} that shows the distribution of raw feature maps in the base set (line 9). Example of a histogram matrix is shown in Figure~\ref{fig:histo_matrix}. 
The next step is to generate raw feature maps, and then a histogram matrix for each driver in the control set (lines 11--20). Then, we obtain the difference between the histogram created for each driver in the control set, and the histogram created for the whole base set (lines 22--26), and summarize these differences to obtain a single matrix that we term the \textit{natural deviation} (line 27). Next, we repeat the same process for the observation set by creating raw feature maps (lines 29--33), and then generating a histogram matrix for driver $d$ (line 34), finally obtaining the difference between the matrix generated for driver $d$ and the base histogram matrix (line 35). %\textcolor{red}{could we add a figure of a histogram matrix where we first mention it - earlier}. 
The output of this algorithm is the difference between the natural deviation and the observed deviation, which we term the \textit{deviation feature map} for driver $d$ (line 36). 

\begin{algorithm}[ht!]
\DontPrintSemicolon
    \KwIn{The base set $\mathrm{B}$, the control set $\mathrm{C}$, and the observation trajectory set $\mathrm{T}_d$}

    $\mathrm{B}_{FM} = [\text{ }]$ \Comment{Set of raw feature maps for the Base set $\mathrm{B}$}\;
    \For{\textit{driver } $k$  \textbf{in}  $\mathrm{B}$}  
    {
        \For{\textit{trajectory } $t$  \textbf{in}  $\mathrm{T}_k$}  
        {
            $fm = \text{build\_raw\_feature\_map }(t)$\;
            $\mathrm{B}_{FM}.\textit{append }(fm)$\;
        }
    }
    
    \Comment{Summarize raw feature maps by creating a histogram matrix for the base set}\;
    $hist\_matrix_\mathrm{B} = \textit{obtain\_histogram\_matrix }(\mathrm{B}_{FM})$\;
    
    \Comment{For each driver in control set, obtain raw features maps and their histogram matrix}\;
    $\mathrm{C}_{HM} = [\text{ }]$ \Comment{Set of histogram matrices for the Control set $\mathrm{C}$}\;
    \For{\textit{driver } $k$  $\in$  $\mathrm{C}$}  
    {
        $k_{FM} = [\text{ }]$\;
        \For{\textit{trajectory } $t$ $\in$ $\mathrm{T}_k$}
        {
            $fm = \text{build\_raw\_feature\_map }(t)$\;
            $k_{FM}.\textit{append }(fm)$\;
        }
        $hist\_matrix_k = \textit{obtain\_histogram\_matrix }(k_{FM})$\;
        $\mathrm{C}_{HM}.\textit{append }(hist\_matrix_k)$\;
    }
    
    \Comment{Obtain the natural deviation by summarizing deviation of the control set from the base set}\;
    $\mathrm{C}_{\textit{diff}} = [\text{ }]$\;
    \For{\textit{hist\_matrix } $hm$ $\in$ $\mathrm{C}_{HM}$}
    {
        $\textit{diff} = \textit{obtain\_difference }(hist\_matrix_\mathrm{B}, hm)$\;
        $\mathrm{C}_{\textit{diff}}.\textit{append }(\textit{diff})$\;
    }
    $\textit{natural\_deviation} = \textit{summarize }(\mathrm{C}_{\textit{diff}})$ \;
    
    \Comment{Generate the deviation feature map for driver $d$, based on the natural deviation}\;
    $d_{FM} = [\text{ }]$ \Comment{Set of raw feature maps for the observation trajectory set}\;
    \For{\textit{trajectory } $t$  $\in$  $\mathrm{T}_d$}  
    {
        $fm = \text{build\_raw\_feature\_map }(t)$\;
        $d_{FM}.\textit{append }(fm)$\;
    }
    $hist\_matrix_d = \textit{obtain\_histogram\_matrix }(d_{FM})$\;
    $\textit{observed\_deviation}_d = \textit{obtain\_difference }(hist\_matrix_\mathrm{B}, hist\_matrix_d)$ \;
    $\textit{deviation\_feature\_map}_d = \textit{difference}(\textit{observed\_deviation}_d, \textit{natural\_deviation})$ \;
    
    \KwOut{$\textit{deviation\_feature\_map}_d$} 
\caption{Generating Deviation Feature Map}
\label{algo:dev_feature_map}
\end{algorithm}

The functions used in Algorithm~\ref{algo:dev_feature_map} are described below. 
\begin{itemize}[leftmargin=0.5cm]
    \item Function \textit{build\_raw\_feature\_map()}: This function creates a raw feature map for an input trajectory. 
    \item Function \textit{obtain\_histogram\_matrix()}: Given a set of raw feature maps, we summarize them by obtaining a histogram that shows the distribution of values in each cell. 
    \item Function \textit{obtain\_difference()}: This function obtains deviation from the base histogram matrix.  %\textcolor{red}{how is this difference computed? Can a figure or example be given?}. 
    As the goal is to obtain difference between two distributions, we can employ any reasonable method to obtain such deviation. Examples of possible solutions are \textit{Kullback–Leibler divergence} \cite{kullback1951information}, \textit{total variation} \cite{total_variation}, and \textit{Hellinger distance} \cite{hellinger1909neue}. %Note that this function uses two matrices as input, where each matrix comprises a set of probability distributions (i.e., one histogram per cell). The final deviation is not just a single value, but a two dimensional matrix where each cell contains the deviation between the probability distributions for the corresponding cells in the input matrices. 
    \item Function \textit{summarize()}: This function summarizes deviation matrices of drivers in the control set, using operations such as summation, average, etc. 
    \item Function \textit{difference()}: This function measures the difference between the natural deviation (i.e., output of the \textit{summarize} function), and the observed deviation for driver $d$, to build a deviation feature map for driver $d$. 
\end{itemize}

One example of deviation feature map that is generated for a driver in the observation set is shown in Figure~\ref{fig:raw_dev_feature_maps}-b. The number of deviation feature maps to represent a driver depends on the number of raw feature maps to represent each trajectory. Further details on various types of raw feature maps used in this work are provided in Section~\ref{sec:feature_maps_for_test}. 

\subsection{Risk Cohort Classifier}
As shown in Figure~\ref{fig:risk_prediction}, the main outcome of the modeling process is a risk cohort prediction classifier. We describe the different requirements and steps of building such a classifier in the following sub-sections. 

\subsubsection{Accident and Citation Records}
This set contains traffic accident and citation records for each driver in the modeling set. In terms of accidents, we only consider the \textit{at-fault} cases. For each driver, we only consider their records in the last five years. %This is because insurance companies usually consider the history of the last five years in the underwriting process. 

\subsubsection{Identifying Risk Cohorts}
%This step aims to refine weak risk labels using contextualized-telematics data and a data-driven process. 
We employ Algorithm~\ref{algo:identify_risk_cohorts} to identify risk cohorts for drivers in the modeling set. Each driver is represented by a set of $\mathcal{D}$ deviation maps. We use each of the deviation maps to cluster the drivers (line 3). After each clustering, we perform a post-processing step to associate each cluster with a risk cohort label (line 4). To do this, we employ the partial (or weak) ground truth risk label data (i.e., traffic conviction records), obtain the average number of records in each cluster, and then associate each cluster with a categorical label (e.g., low-risk and high-risk)\footnote{For instance, if we define our labels as low-risk and high-risk, then the cluster with smaller average of conviction records per driver will be labeled as low-risk, and the other cluster will be labeled as high-risk.}. 
Then, for each driver, we update their label set by adding the newly obtained cohort label (lines 5--7). Finally, we find the most frequent label for each driver given a confidence coefficient $\mathcal{C}$ (lines 9--16) as a threshold. Based on this threshold, we identify those drivers who have been confidently assigned to a specific risk cohort. As an example, if a driver is assigned to the \textit{low-risk} cohort (say) $80\%$ of the time, we label her as a \textit{low-risk} driver. If a driver found to be not confidently assigned to any of the cohorts, then her final label would be $Null$. The output of this algorithm would be a subset of drivers in the modeling set with a valid final risk cohort label. 

\begin{algorithm}[ht!]
\DontPrintSemicolon
    \KwIn{Modeling set drivers $\mathcal{M}$, Deviation feature maps $\mathcal{D}$, number of cohorts $k$, traffic conviction records $Recs$, and labeling confidence $\mathcal{C}$.}
    $\textit{driver\_to\_cohort\_label} = \{\}$\;
    \For{\textit{deviation\_map} $D \in \mathcal{D}$}
    {
        $res = \textit{do\_clustering }(\mathcal{M}_D, k)$ \Comment{\small $\mathcal{M}_D$ represents the modeling set drivers by deviation map $D$}\;
        $\textit{cohort\_label} = \textit{clustering\_label\_to\_risk\_cohort\_label }(res, Recs)$
        
        \For{\textit{driver} $d \in \mathcal{M}$}
        {
            $\textit{driver\_to\_cohort\_label }[d].\textit{append }(\textit{cohort\_label }[d])$\;
        }
    }
    
    $\textit{final\_labeled\_set} = \{\}$\;
    \For{\textit{driver} $d \in \mathcal{M}$}
    {
        $labels = \textit{driver\_to\_cohort\_label }[d]$ \Comment{\small Labels assigned to driver $d$ by clustering tasks} \;
        $\textit{final\_label} = \textit{find\_frequent\_label }(labels, \mathcal{C})$ \;
        
        \If{\textit{final\_label} $\neq$ \textit{Null}}
        {
            $\textit{final\_labeled\_set }[d] = \textit{final\_label}$\;
        }
    }
    
    \KwOut{\textit{final\_labeled\_set}} 
\caption{Identifying Risk Cohorts}
\label{algo:identify_risk_cohorts}
\end{algorithm}

While any clustering algorithm may be used in the $\textit{do\_clustering()}$ function, \textit{K-means} is a reasonable choice given that the number of clusters must be specified beforehand. In terms of number of clusters, we can pick any value greater than 1, depending on the driving context and application requirements. In this work, we use a value of 2 to test our framework. %Also, we determine the labeling confidence $\mathcal{C}$ based on the number deviation maps to represent each driver, mainly because this value determines the number of potential labels for each driver. For instance, if we use 10 deviation maps for each driver, then we can choose any confidence threshold from the following set: $\langle0.5, 0.6, 0.7, 0.8, 0.9 \rangle$. 

\subsubsection{Constructing the Risk Cohort Classifier}
The last step of the modeling process is to build a classifier that uses deviation maps as input to predict the risk cohort label (or the refined risk label). The input of this part is the set of labeled drivers and their deviation maps obtained from the previous step. Any off-the-shelf classifier can potentially be used to predict the risk cohort. %The choice of the classifier depends on the size of the data and the complexity of the prediction task.  

\vspace{5pt}
It is worth noting again that \textit{refining risk labels} is an important aspect of our framework. To demonstrate its importance, suppose that we build a classifier to predict driving risk using telematics data as input and using past traffic records as risk labels. Although this seems a straightforward setting, it may not result in any satisfactory outcomes, due to \textit{unreliability} of the raw risk labels. An important practical observation here is that if someone only has records of citation or accident it does not mean that she is a risky driver; and only because someone does not have such records, it does not necessarily mean that she is a safe driver. Consequently, we need a process to refine our labels prior to building a risk cohort classifier. %Our label refinement process primarily uses telematics data, with traffic citation and accident records only as \textit{partial} ground-truth data. 

\subsection{The Prediction Process}
In order to employ the proposed framework for the real-world applications, we describe a \textit{prediction process}. Given a set of trajectories for a new or an unseen driver, the important steps of this process are (1) Creating raw feature maps, (2) Creating deviation feature maps, and (3): Risk prediction using risk cohort classifier.

\section{Datasets and Data Representation}
In this section we first describe our datasets, and then describe telematics data representation via features maps. 

\subsection{Datasets}
\label{sec:dataset}
We employed multiple and extensive datasets to build and evaluate the proposed risk prediction framework. Data were collected from the {\em CAN-bus}\footnote{Controller Area Network (or CAN-bus) is a communication mechanism to transfer a variety of information related to the operation of a vehicle.} and other sensors including GPS, accelerometer, and magnetometer, and for five cities\footnote{Atlanta (GA), Columbus (OH), Memphis (TN), Philadelphia (PA), and Pittsburgh (PA).}. In addition to trajectory information, we have the demographic data for drivers and also their history of traffic accidents or citation records. In building the framework, we divided our data into multiple sets, as described below. 
%\textcolor{red}{Are their more standard names for Modelling, Reference and Held-Out sets? Training, Validation, Testing?}
\begin{itemize}[leftmargin=0.5cm]
    \item \textbf{Modeling Set}: This is the set that we use for the modeling process to create the risk cohort prediction classifier. In this set, we have about 4,000 drivers, none of whom shared their vehicle with any other driver, and they owned exactly one vehicle in our data. For each driver, we sampled exactly 100 distinct trajectories. 
    \item \textbf{Reference Set}: This is the set of ``presumably safe drivers'', with no records of accidents or traffic citations from 2010 to 2019. In this set, we have 5,000 vehicles. Each vehicle could have been shared with more than one driver, but all drivers who shared the same vehicle satisfied the previous constraint. We sampled 100 distinct trajectories for each vehicle in this set. 
    \item \textbf{Held-out Set}: This is a smaller set to evaluate the prediction process. It includes 250 drivers who owned exactly one vehicle, and no vehicle was shared between the drivers. 
\end{itemize}

To form the partial ground-truth labels for drivers in the modeling and the held-out sets, we collected their traffic citation and at-fault accident records from 2014 to 2019. 

\subsection{Data Representation by Raw Feature Maps}
\label{sec:feature_maps_for_test}
We represent a trajectory using 22 different feature maps, and categorize them in terms of six groups described in Table~\ref{tab:feat_maps}. A few notes regarding definition of raw feature maps:

\begin{table*}[ht!]
    \centering
    \caption{Description of Raw Feature Maps (attributes and contexts)}
    \begin{tabular}{c|c|c|c|c|c}
        \rowcolor{Blue}\textbf{Group} & \textbf{Feature Map} & \textbf{Attributes} & \textbf{Context} & \textbf{$1^{st}$ Attr Range} & \textbf{$2^{nd}$ Attr Range} \\
        \hline
        \hline
        \multirow{3}{*}{\textbf{G1}} & T1 & Speed, Acceleration & N/A & [5, 25] mph & [-2.5, 2.5] $m/s^2$ \\
                            & T2 & Speed, Acceleration & N/A & [25, 45] mph & [-2, 2] $m/s^2$ \\
                            & T3 & Speed, Acceleration & N/A & [45, 80] mph & [-1, 1] $m/s^2$ \\
        \hline
        \multirow{5}{*}{\textbf{G2}} & T4 & Speed, Acceleration & Urban roads & [5, 25] mph & [-2.5, 2.5] $m/s^2$ \\
                            & T5 & Speed, Acceleration & Urban roads & [25, 45] mph & [-2, 2] $m/s^2$ \\
                            & T6 & Speed, Acceleration & Highway roads & [5, 25] mph & [-2.5, 2.5] $m/s^2$ \\
                            & T7 & Speed, Acceleration & Highways roads & [25, 45] mph & [-2, 2] $m/s^2$ \\
                            & T8 & Speed, Acceleration & Highways roads & [45, 80] mph & [-1, 1] $m/s^2$ \\
        \hline
        \multirow{6}{*}{\textbf{G3}} & T9 & Speed, Acceleration & Day time & [5, 25] mph & [-2.5, 2.5] $m/s^2$ \\
                            & T10 & Speed, Acceleration & Day time & [25, 45] mph & [-2, 2] $m/s^2$ \\
                            & T11 & Speed, Acceleration & Day time & [45, 80] mph & [-1, 1] $m/s^2$ \\
                            & T12 & Speed, Acceleration & Night time & [5, 25] mph & [-2.5, 2.5] $m/s^2$ \\
                            & T13 & Speed, Acceleration & Night time & [25, 45] mph & [-2, 2] $m/s^2$ \\
                            & T14 & Speed, Acceleration & Night time & [45, 80] mph & [-1, 1] $m/s^2$ \\
        \hline
        \multirow{3}{*}{\textbf{G4}} & T15 & Speed, Acceleration & Turn segments & [5, 25] mph & [-3, 3] $m/s^2$ \\
                            & T16 & Speed, Acceleration & Turn segments & [25, 45] mph & [-3.5, 3.5] $m/s^2$ \\
                            & T17 & Speed, Acceleration & Turn segments & [45, 80] mph & [-2, 2] $m/s^2$ \\
        \hline
        \multirow{3}{*}{\textbf{G5}} & T18 & Speed, AngleChange & Turn segments & [5, 25] mph & Up to 20 degrees \\
                            & T19 & Speed, AngleChange & Turn segments & [25, 45] mph & Up to 6 degrees \\
                            & T20 & Speed, AngleChange & Turn segments & [45, 80] mph & Up to 4 degrees \\
        \hline
        \multirow{2}{*}{\textbf{G6}} & T21 & Acceleration, AngleChange & Turn segments & [-2, 2] $m/s^2$ & Up to 7 degrees \\
                            & T22 & AngularSpeed, AngleChange & Turn segments & [0, 0.025] rph & Up to 15 degrees \\
    \end{tabular}
    \label{tab:feat_maps}
\end{table*}

\begin{itemize}[leftmargin=0.5cm]
    \item Attributes' Order: The order of attributes in each feature map suggests that the first attributes is used in y-axis and the second attribute in x-axis (see section~\ref{subsub:raw_feature_map}). 
    \item Context: Shows the context that is used to generate a feature map (if any). Based on this, feature maps in group G1 (see Table~\ref{tab:feat_maps}) are context-agnostic. 
    \item Urban Roads: We label a road segment as an \textit{urban road} if its speed-limit falls in the interval [0, 50] mph. We obtained speed-limit data by using the Graphhopper Map-matching Library \cite{graphhopper}. To add more details, we map-matched each trajectory using location and time data, and obtained speed-limit information for each segment that constituted a trajectory.  
    \item Highway Roads: We label a road segment as a \textit{highway} (or high-speed) road, if its speed-limit falls in the interval [50, 90] mph. Please note that the definition of road type, as we defined in this work (i.e., urban versus highway), is a practical choice after trying a few settings and analyzing their impact on final risk prediction results\footnote{Interested readers may try other settings to see their impacts}. 
    \item Day Time: We used \textit{civil twilight system}\footnote{See more at \url{https://en.wikipedia.org/wiki/Twilight\#Civil\_twilight}} to annotate each trajectory based on its start time. In this sense, ``day time'' indicates a trajectory was generated during the day time based on the civil twilight system. We obtained daylight information from \cite{timeanddate} to annotate each trajectory. 
    \item Night Time: This contextual concept indicates that a trajectory was generated during the night time based on the civil twilight system. 
    \item Turn Segments: This contextual concept means representing trajectory attributes for turn segments that we extract from a trajectory. We use a fairly accurate turn detection algorithm that is proposed in \cite{moosavinejaddaryakenari2020telematics} (see Section 6.4.4) to extract turn segments. 
    \item AngleChange: This is a basic feature extracted from a trajectory that shows the change of \textit{angle} between two consecutive observations. Here, we obtain angle based on three location points (e.g., A, B, and C) and using Equation~\ref{eq:angle}. Thus, angle shows orientation of a moving vehicle, and change in angle implies change in orientation. 
    
    \begin{equation}
        \theta = \arccos \left( \frac{\vec{AB}\cdot \vec{BC}}{ \|\vec{AB}\| \, \|\vec{BC}\|}\right)  
        \label{eq:angle}
    \end{equation}
    
    \item AngularSpeed: This is another basic feature that measures angular displacement per a unit of time. We used Equation~\ref{eq:angl_speed} to measure angular speed in radian per hour (rph). Here $H(.)$ calculates haversine distance in terms of radian between two coordinates $P_1$ and $P_2$, and $r$ is the Earth radius (i.e., $6.3781\times 10^6$ meters). 
    
    \begin{equation}
        \textit{Angular\_Speed($P_1$, $P_2$)} = \textit{H}(P_1, P_2)\times \frac{3600}{r}
        \label{eq:angl_speed}
    \end{equation}
    
\end{itemize}
\section{Experiments and Results}
\label{sec:results}
In this section, we evaluate our risk prediction framework based on real-world data. First, we describe a set of raw feature maps that we created for each trajectory. Then, we present details of our experiments and results. 

%\textcolor{red}{This section should be about the experiments, not describing the methods. Should Raw Feature Maps move to the section earlier - say the Data subsection - since it is a description of the method?}

%\input{tex/feature_maps.tex}

\subsection{Risk Cohort Prediction Results}
After representing a trajectory using the 22 different raw feature maps, we created the risk cohort classifier from the modeling and reference sets (see Section~\ref{sec:method} for details). We examined different combinations of raw feature maps, and the set of all feature maps to represent trajectories. Next we describe several settings before presenting the results.  

\begin{itemize}[leftmargin=0.5cm]
    \renewcommand{\labelitemi}{\tiny$\blacksquare$} 
    
    \item \textbf{Choice of Deviation Metric}: To obtain the deviation between two probability distributions in Algorithm~\ref{algo:dev_feature_map}, we may employ any deviation calculation method. However, for the sake of simplicity, we use \textit{Hellinger distance} \cite{hellinger1909neue} for discrete distributions (Equation~\ref{eq:hellinger}):
    \begin{equation}
        \small \textit{Hel\_Dist}(A, B) = \frac{1}{2} \sqrt{\sum_{i=1}^{N} \big(\sqrt{A[i]} - \sqrt{B[i]}\big)^2}    
        \label{eq:hellinger}
    \end{equation}
    Here $A$ and $B$ are two vectors. 
    
    \item \textbf{Labeling Confidence $\mathcal{C}$}: The labeling confidence in Algorithm~\ref{algo:identify_risk_cohorts} determines the final labeled set of drivers. We determine this value with respect to the number of deviation maps that we use to represent each driver\footnote{Which depends on the number of raw feature maps to represent each trajectory}, and also the number of risk cohorts. %For example, if we represent each driver using 10 different deviation maps and the objective is to find two risk cohorts, then we use the following confidence set: $\{0.6, 0.7, 0.8, 0.9, 1.0\}$, where $\mathcal{C}=0.6$ means that $60\%$ of labels found for a driver must belong in the same cohort (say low-risk cohort). Given a confidence set $\mathrm{C}$, we choose a value $\mathcal{C} \in \mathrm{C}$ that provides the best prediction results on the held-out set.
    
    \item \textbf{Number of Cohorts}: One of the advantages of this framework is that we may define any number of cohorts. In this work, we used two cohorts, which we referred to as \textit{low-risk} and \textit{high-risk}. 

    \item \textbf{Clustering Algorithm}: The choice of the clustering algorithm to be employed by Algorithm~\ref{algo:identify_risk_cohorts} is another customizable part of this framework. We use K-Means clustering in this work, but any other method that allows the user to set the number of clusters could be suitable. 
    
    \item \textbf{Classification Method}: We used two different models for classification, Gradient Boosting Classifier (GBC) and Convolutional Neural Network (CNN). 
        \begin{itemize} [leftmargin=0.6cm]
            \item GBC: gradient boosting is a strong general-purpose classifier, which usually performs very well on binary or multi-label classification tasks. The input to GBC should be in the form of a vector, which we create by converting deviation feature maps to a vector. Each vector represents one driver, and the risk label for each driver is obtained from the clustering (see Section~\ref{sec:method}). We used grid-search to obtain optimal parameters of GBC classifier using a 5-fold cross validation over the modeling set\footnote{Optimal parameters were found as: $learning\_rate = 0.1$, $num\_estimators=300$, and $max\_depth=5$.}.
            \item CNN: given that our input data is in the form matrix representation, where each matrix represents one feature map, the use of convolutional neural network seems to be a reasonable choice for this problem. Besides the type and representation of input data, the use of CNN may further help to extract latent information from input, and use them to better classify drivers. After hyper-parameter tuning via grid search over the modeling set, the final CNN model includes two convolutional layers (each with 16 filters, kernel size $5\times5$, and stride=1) with max pooling on top of convolutional output (each with pool size $2\times2$ and stride=2), followed by three dense layers. All layers use $ReLU$ as activation function, except the last dense layer which uses \textit{softmax}. Lastly, Root Mean Square Propagation optimizer \cite{tieleman2012divide} (RMSProp) with initial learning rate $0.001$ is used as the optimization method in training.
        \end{itemize}
    
    \item \textbf{Baseline Method}: As baseline, we choose a variation of our proposed solution that employs raw feature maps, and uses the original risk labels to build the risk classifier. In other words, we seek to analyze the impact of two components by this choice: 1) transforming raw feature maps to deviation maps, 2) refining risk labels by clustering. Based on this approach, we represent each driver by an average feature map for each type listed in Table~\ref{tab:feat_maps}. Also, we use a binary label based on drivers' past records to label them, thus a driver with no record would be labeled by 0 (i.e., safe) or 1 otherwise. 
\end{itemize}

\begin{figure*}[t]
    \small
    \centering
    \includegraphics[scale=0.5]{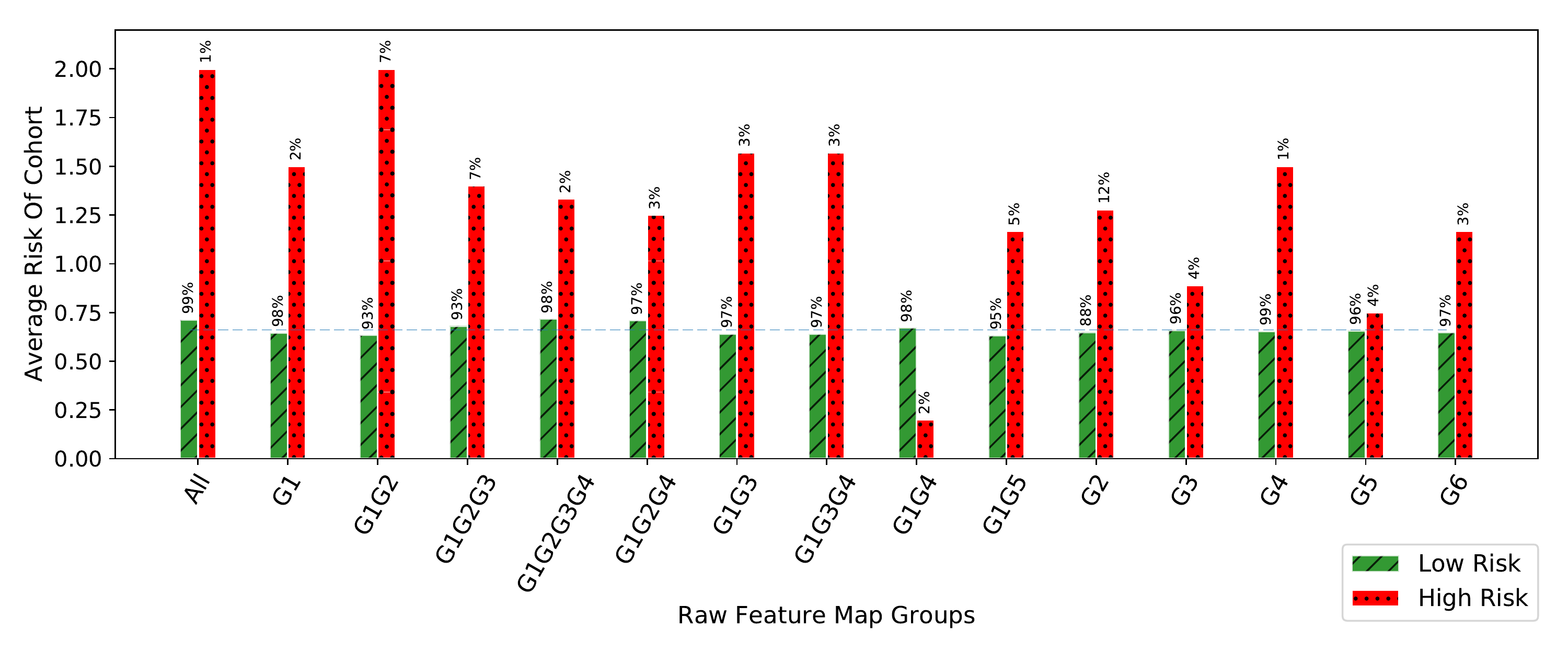} \hspace{-10pt}
    \caption{Baseline risk cohort prediction results for drivers in the held-out set. X-axis shows the combination of raw feature maps, and Y-axis shows the average number of traffic records in each cohort, and blue dashed lines show the average number of records in the held-out set (\% numbers on bars show relative size of each cohort).}
    \label{fig:risk_prediction_results_baseline}
\end{figure*}

\subsection{Risk Prediction Results on the Held-out Set}
\label{subsec:res_overall}

We used different combinations of raw feature maps to build our risk prediction pipeline and predict the binary risk cohort label for drivers in the held-out set (i.e., 250 drivers). For the drivers in each cohort, we represent the average cohort risk in terms of the average number of traffic citations or accident records in the past five years. Results are shown in Tables~\ref{tab:risk_prediction_results_gbc} and \ref{tab:risk_prediction_results_cnn} (for the proposed framework) and Figure~\ref{fig:risk_prediction_results_baseline} (for the baseline method). 

\begin{table*}[ht!]
    % \footnotesize 
    \centering
    \caption{Risk cohort prediction results for held-out set using \textbf{GBC} model. Here we have two risk cohorts: \textit{low-risk} and \textit{high-risk}. The proportion of drivers with no traffic accident or citation records is shown by \textit{no records \%} for each cohort. Average risk of each cohort shows the average number of records per driver in each cohort. Rows colored by light green show the ones that provided the best results according to our problem formulation in Section~\ref{sec:problem}.}% \vspace{-10pt}
    \hspace*{-20pt}
    \begin{tabular}{c||c|c||c|c||c|c}
        \textbf{\begin{tabular}{@{}c@{}} Feature \\ Groups \end{tabular}} & \textbf{\begin{tabular}{@{}c@{}} Low-Risk \\ Drivers \%\end{tabular}} & \textbf{\begin{tabular}{@{}c@{}} High-Risk \\ Drivers \%\end{tabular}} & \textbf{\begin{tabular}{@{}c@{}} Low-Risk \\ no records \%\end{tabular}} & \textbf{\begin{tabular}{@{}c@{}} High-Risk \\ no records \%\end{tabular}} & \textbf{\begin{tabular}{@{}c@{}} Avg. risk of \\ Low-Risk\end{tabular}} & \textbf{\begin{tabular}{@{}c@{}} Avg. risk of \\ High-Risk \end{tabular}}\\
        \hline
        G1 & 43\% & 57\% & 64.95\% & 55.12\% & 0.526 & 0.764 \\
        G2 & 38\% & 62\% & 63.79\% & 51.06\% & 0.569 & 0.819 \\
        G3 & 33\% & 67\% & 67.12\% & 54.48\% & 0.452 & 0.779 \\
        G4 & 41\% & 59\% & 58.24\% & 60.15\% & 0.659 & 0.662 \\
        \rowcolor{LightGreen}
        G5 & 51\% & 49\% & 67.83\% & 50.46\% & 0.443 & 0.890 \\
        G6 & 43\% & 57\% & 64.95\% & 55.12\% & 0.557 & 0.740 \\
        G1G2 & 32\% & 68\% & 65.31\% & 51.46\% & 0.449 & 0.854 \\
        G1G3 & 41\% & 59\% & 63.33\% & 55.47\% & 0.522 & 0.773 \\
        G1G4 & 48\% & 52\% & 61.11\% & 57.76\% & 0.556 & 0.759 \\
        G1G5 & 48\% & 52\% & 65.74\% & 53.45\% & 0.500 & 0.810 \\
        \rowcolor{LightGreen}
        G1G2G3 & 25\% & 75\% & 75.68\% & 48.65\% & 0.297 & 0.874 \\
        G1G2G4 & 32\% & 68\% & 68.75\% & 50.00\% & 0.500 & 0.827 \\
        G1G3G4 & 43\% & 57\% & 65.96\% & 53.23\% & 0.489 & 0.806 \\
        G1G2G3G4 & 24\% & 76\% & 75.00\% & 49.11\% & 0.306 & 0.866 \\
        All & 26\% & 74\% & 73.68\% & 49.09\% & 0.316 & 0.873 \\
    \end{tabular}
    \label{tab:risk_prediction_results_gbc}
\end{table*}

\begin{table*}[ht!]
    % \footnotesize 
    \centering
    \caption{Risk cohort prediction results for held-out set using \textbf{CNN} model}% \vspace{-10pt}
    \hspace*{-20pt}
    \begin{tabular}{c||c|c||c|c||c|c}
        \textbf{\begin{tabular}{@{}c@{}} Feature \\ Groups \end{tabular}} & \textbf{\begin{tabular}{@{}c@{}} Low-Risk \\ Drivers \% \end{tabular}} & \textbf{\begin{tabular}{@{}c@{}} High-Risk \\ Drivers \% \end{tabular}} & \textbf{\begin{tabular}{@{}c@{}} Low-Risk \\ no records \%\end{tabular}} & \textbf{\begin{tabular}{@{}c@{}} High-Risk \\ no records \%\end{tabular}} & \textbf{\begin{tabular}{@{}c@{}} Avg. risk of \\ Low-Risk\end{tabular}} & \textbf{\begin{tabular}{@{}c@{}} Avg. risk of \\ High-Risk \end{tabular}}\\
        \hline
        G1 & 70\% & 30\% & 63.46\% & 50.00\% & 0.551 & 0.912 \\
        G2 & 36\% & 64\% & 70.37\% & 47.96\% & 0.426 & 0.888 \\
        G3 & 42\% & 58\% & 67.03\% & 52.76\% & 0.473 & 0.811 \\
        G4 & 74\% & 26\% & 61.45\% & 53.45\% & 0.596 & 0.845 \\
        \rowcolor{LightGreen}
        G5 & 48\% & 52\% & 71.03\% & 48.72\% & 0.364 & 0.932 \\
        G6 & 44\% & 56\% & 61.62\% & 57.60\% & 0.606 & 0.704 \\
        G1G2 & 39\% & 61\% & 65.00\% & 50.00\% & 0.483 & 0.880 \\
        G1G3 & 48\% & 52\% & 66.67\% & 51.33\% & 0.476 & 0.850 \\
        G1G4 & 49\% & 51\% & 64.55\% & 54.39\% & 0.509 & 0.807 \\
        G1G5 & 57\% & 43\% & 66.41\% & 50.00\% & 0.477 & 0.906 \\
        \rowcolor{LightGreen}
        G1G2G3 & 28\% & 72\% & 76.19\% & 47.17\% & 0.286 & 0.906 \\
        G1G2G4 & 37\% & 63\% & 67.86\% & 48.96\% & 0.464 & 0.875 \\
        G1G3G4 & 27\% & 73\% & 69.49\% & 54.72\% & 0.390 & 0.774 \\
        G1G2G3G4 & 28\% & 72\% & 71.43\% & 49.06\% & 0.381 & 0.868 \\
        All & 28\% & 72\% & 73.17\% & 48.60\% & 0.317 & 0.888 \\
    \end{tabular}
    \label{tab:risk_prediction_results_cnn}
\end{table*}

We can summarize the main takeaways as follows. 

\begin{figure*}[h]
    \hspace{-.5cm}
    \minipage{0.25\textwidth}
        \small
        \centering
        \includegraphics[width=\linewidth]{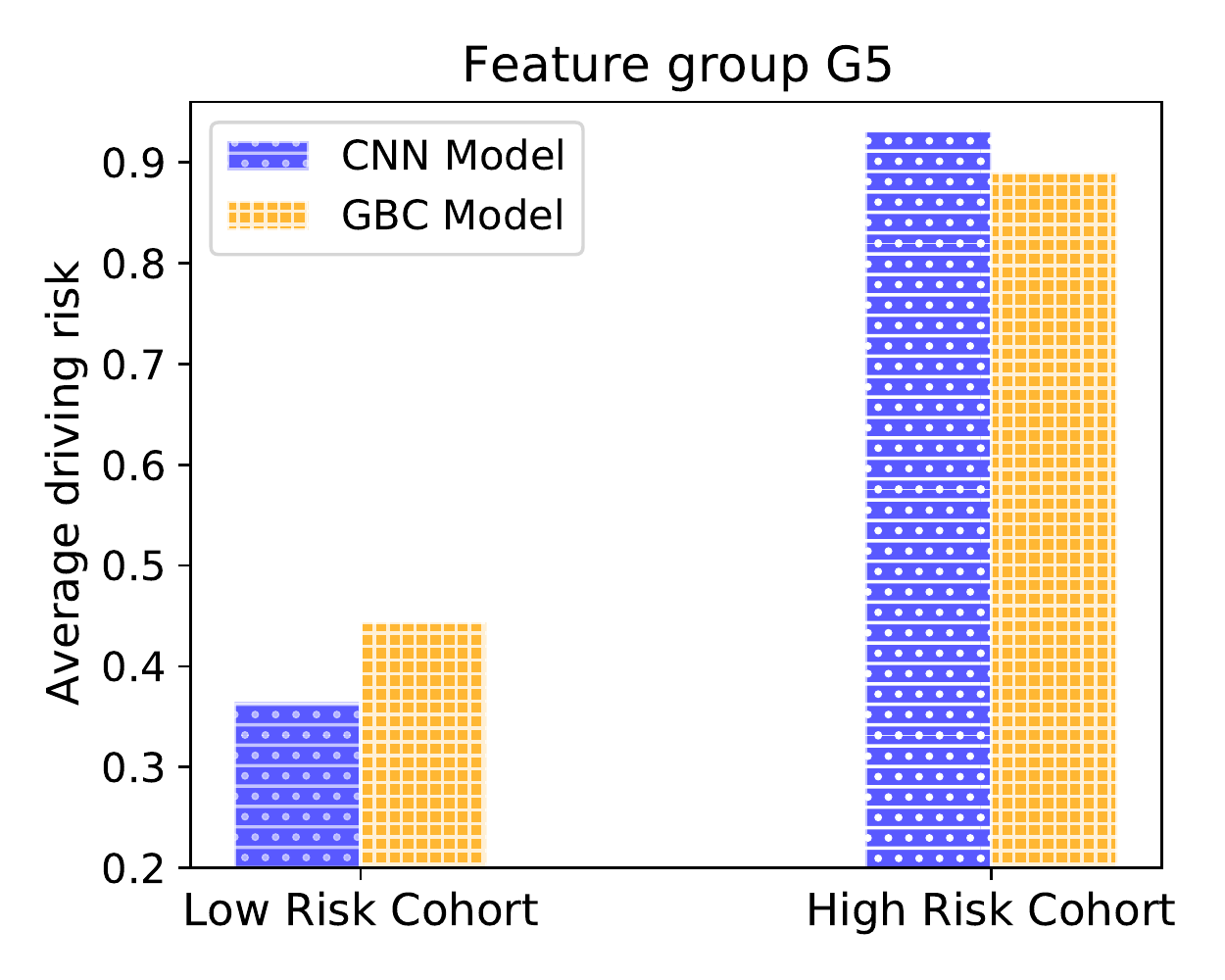}
        (a) Average risk using feature group G5
    \endminipage
    \minipage{0.25\textwidth}
        \small
        \centering
        \includegraphics[width=\linewidth]{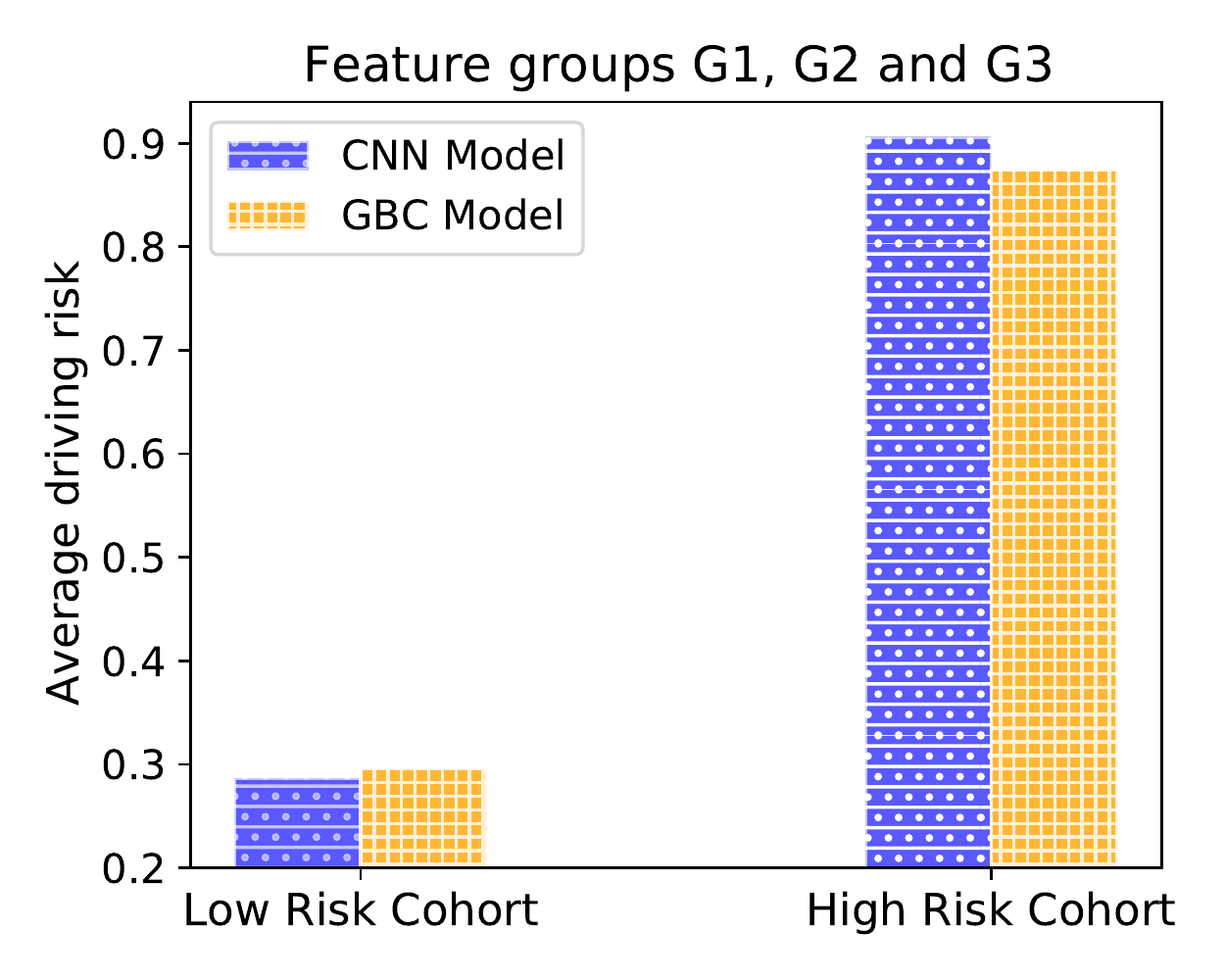}
        (b) Average risk using G1, G2, G3
    \endminipage
    \minipage{0.25\textwidth}
        \small
        \centering
        \includegraphics[width=\linewidth]{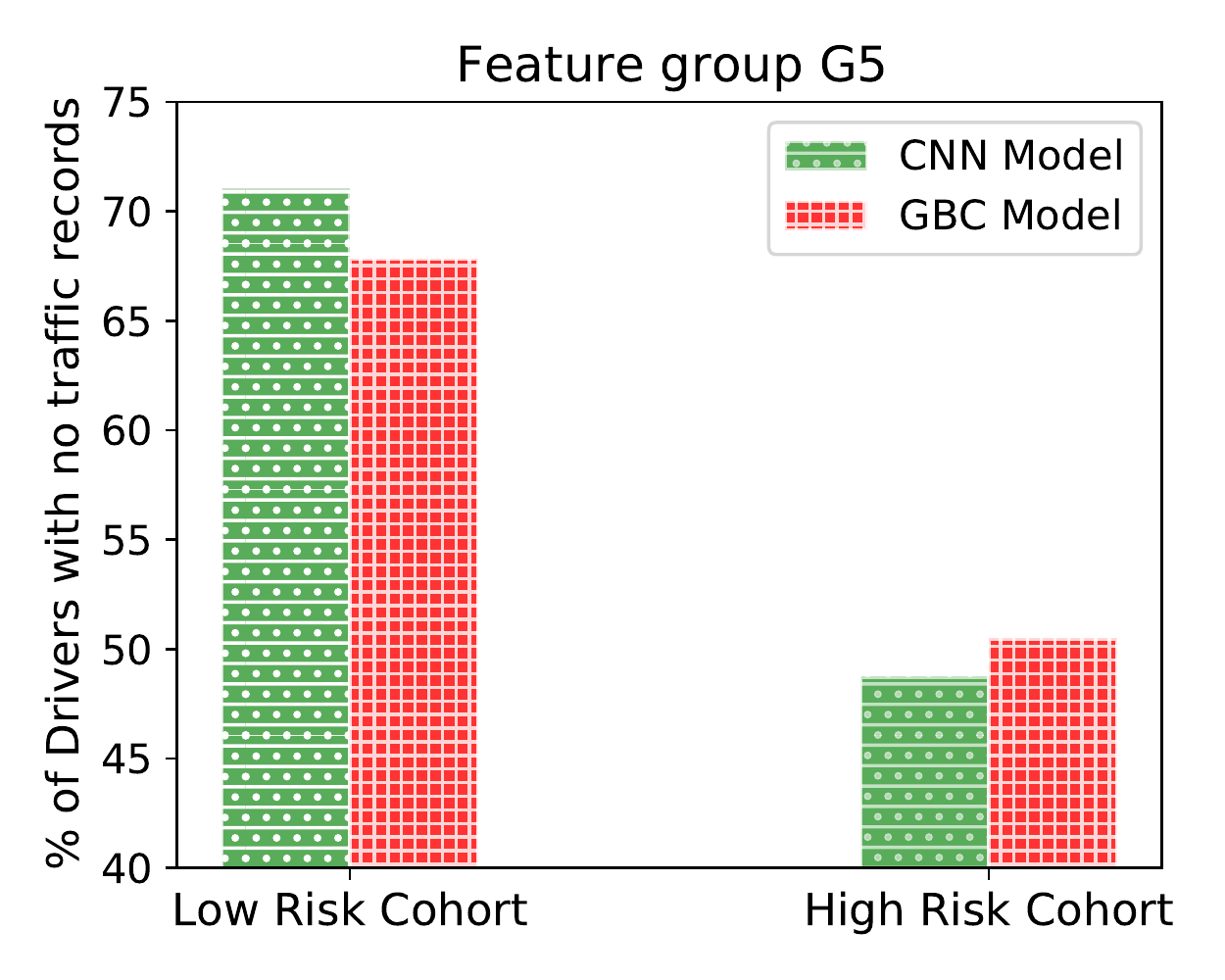}
        (c) \% of drivers with no records using G5
    \endminipage
    \minipage{0.25\textwidth}
        \small
        \centering
        \includegraphics[width=\linewidth]{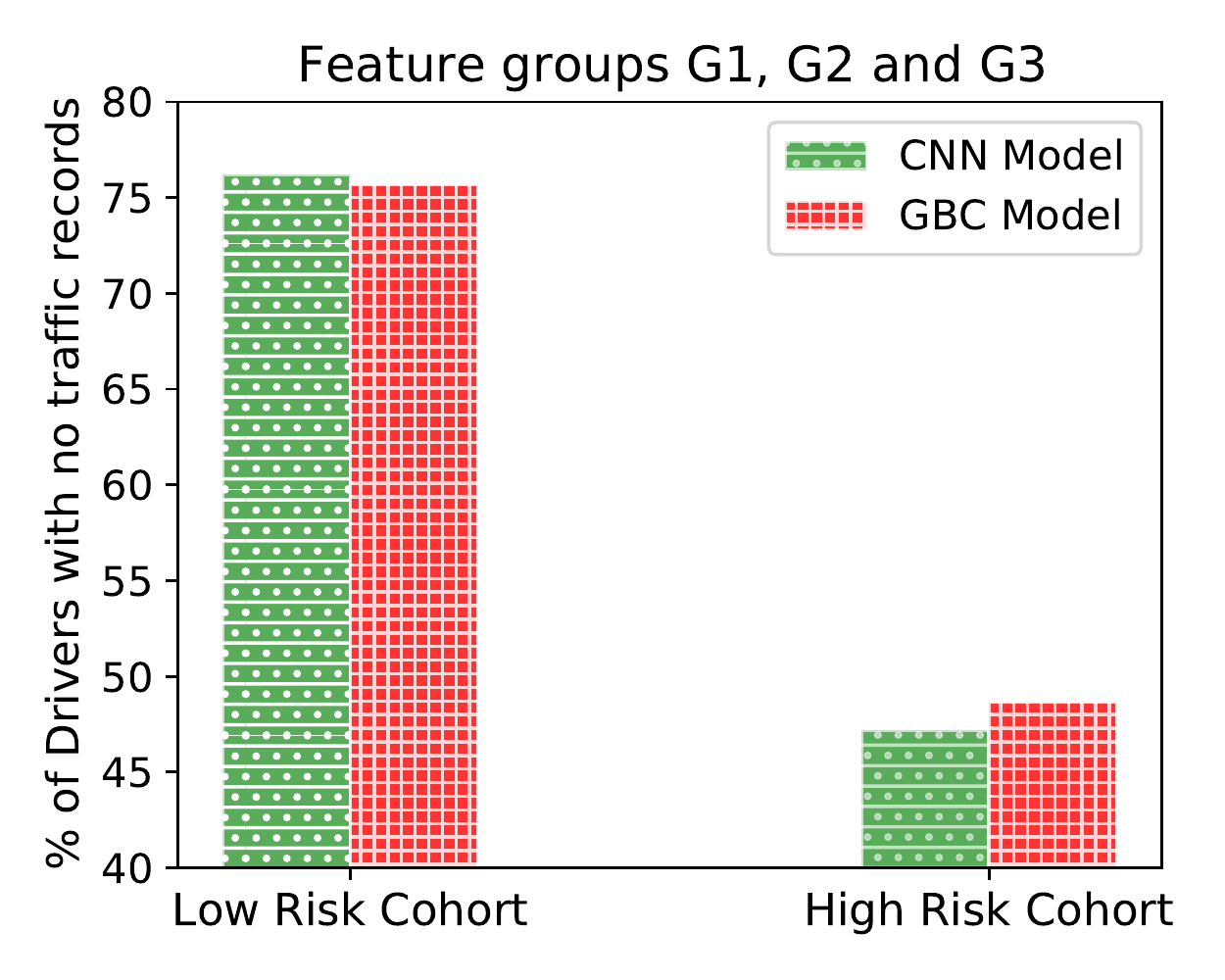}
        (d) \% of drivers with no records using G1, G2, G3
    \endminipage
    \hspace{-.5cm}
    \minipage{\textwidth}
        \vspace{4pt}
        \centering
        \caption{Comparing GBC and CNN models based on average risk score in low- and high-risk cohorts (a and b), and \% of drivers with no citation or accident records in each cohort (c and d). According to these results, the CNN model outperforms the GBC model by finding better cohorts using the two best sets of feature groups.}
        \label{fig:cnn_vs_gbc}
    \endminipage

\end{figure*}

\begin{itemize}[leftmargin=0.5cm]
    \item \textbf{Best combination of feature groups}: The results imply that the features maps T1 to T14 (i.e., the feature groups 1, 2, and 3) provided the best risk cohort classification results, considering the ``difference between average risk scores'' obtained for low- versus high-risk cohorts. It is also interesting to note that about 76\% of drivers in the low-risk cohort do not have any traffic records (i.e., traffic citations or accidents) in the past five years according to Tables~\ref{tab:risk_prediction_results_gbc} and \ref{tab:risk_prediction_results_cnn}. 
    
    \item \textbf{Best single feature group}: The fifth feature group that includes feature maps T18, T19, and T20 provided the best risk cohort prediction results (based on difference in terms of average risk score for low-risk versus high-risk cohort). This group of feature maps are the ones based on speed and angle change, but for turn segments. This is an interesting observation, because \textit{turns} represent an important aspect of ``driving behavior'', and they seem to provide a very strong signal to identify risky drivers. %We also note that about 71\% of drivers in the low-risk cohort do not have any traffic records in the past 5 years, while this percentage for the high-risk cohort is about 49\% based on the CNN model (see Table~\ref{tab:risk_prediction_results_cnn}). Lastly, we see a better ``balance'' in terms of number of drivers in each cohort when comparing it to the best combination of feature groups that discussed above. 
    
    \item \textbf{Impact of deviation maps and risk-relabeling}: The results in Figure~\ref{fig:risk_prediction_results_baseline} clearly highlight the importance of two essential components of our proposal, 1) deviation feature maps, and 2) risk relabeling. None of the combinations of feature maps resulted in any better results than having two very imbalanced cohorts, where the average risk score for the larger cohort is almost the same as the overall average risk score in the held-out set (i.e., the blue dashed line). On a complementary note, these results show that we may not perform a reasonable driving risk prediction only based on raw telematics data and based on the original (i.e., weak or partial) risk labels. Instead, we require proper transformations to make the prediction task a viable one, similar to what we proposed in this paper. 

    \item \textbf{Impact of contextual information: Road type}: Based on comparing feature groups G1 and G2, we notice that including road type information as a contextual factor improves the separation between low- and high-risk drivers based on average risk score and proportion of drivers with no records in each cohort. This highlights the importance of contextual information in determining driving risk.
    
    \item \textbf{Impact of contextual information: Daylight}: When comparing feature groups G1 and G3, we see that adding daylight information (daytime versus night time) has a significant impact on determining driving risk. The inclusion of this information changes the classification of some drivers from low-risk to high-risk, leading to an increase in the proportion of drivers with no past records and a decrease in average risk for the low-risk cohort.
    
    \item \textbf{Impact of contextual information: Road shape}: Road shape encoded as turn maneuvers improved the distribution of drivers between low-risk and high-risk cohorts when represented by speed and angle change (G5), while other combinations (G4 and G6) resulted in worse outcomes compared to G1. This highlights the importance of the choice of data attributes in representing telematics data.
    
    %\item \textbf{Imbalanced Cohorts}: While some of the feature groups resulted in slight or severe imbalanced cohorts (in terms of number of drivers in each cohort), we cannot solely attribute this to a bad outcome, because there is no oracle to tell what the ideal relative cohorts sizes should be. That being said, we may still prefer to see more evenly distributed cohorts. 
    
    \item \textbf{The Better Risk Classifier}: According to our results, the use of CNN model resulted in better risk prediction results. Based on feature group G5, the two classifiers found relatively similar sized low- and high-risk cohorts (see the first highlighted row in Tables~\ref{tab:risk_prediction_results_gbc} and \ref{tab:risk_prediction_results_cnn}). However, we observe a larger gap between the cohorts found by the CNN model when comparing their average risk as well as percentage of drivers with no records (see Figures \ref{fig:cnn_vs_gbc}-a and \ref{fig:cnn_vs_gbc}-c). Further, the same pattern exists when looking into the results by a combination of feature groups G1, G2 and G3 (see Figures \ref{fig:cnn_vs_gbc}-b and \ref{fig:cnn_vs_gbc}-d). We assume the superior results by the CNN model are due to its ability to better capture and encode spatio-temporal patterns when a matrix-shaped input is given. 

\end{itemize}

It it worth noting that in most cases, the proportion of drivers with no traffic accidents or citations was higher in the low-risk group than the high-risk group. However, some low-risk drivers had records and vice versa, indicating that having or not having records does not determine risk level. Nevertheless, most drivers with records may be considered high-risk. 
\section{Summary and Conclusions}
\label{sec:conclusion}
In this paper, we introduced a novel driving risk prediction framework that utilizes telematics and contextual data. The framework comprises a modeling and a prediction process, with the former building a risk cohort classifier based on contextualized telematics data. Key steps in the modeling process include constructing contextualized views of trajectories, generating deviation-based views for drivers, refining weak risk labels using a data-driven approach, and building a risk cohort classifier using contextualized telematics data and the improved risk labels. The prediction process employs the trained risk cohort classifier to predict driving risk cohorts using contextualized telematics data. An important aspect of this framework is its risk label refinement step, which employs a novel data-driven approach to improve the quality of the risk labels, facilitating the training of the risk cohort classifier. 

Regarding contextual data, we utilized road type, road shape, and daylight information. However, incorporating other data such as weather and traffic could enhance the framework by creating more feature and deviation maps to represent drivers. Additionally, the framework allows any contextualized view of a trajectory to be used as input and permits adjusting the number of risk cohorts for different regions. Furthermore, the clustering algorithm, classification method, and deviation calculation formulation can be selected based on specific needs. 

While the telematics data used in this research is primarily derived from the CAN-bus and GPS sensor, other sensors such as the accelerometer can also provide valuable information about driving behavior. High-quality accelerometer data collected on high frequency rates (e.g., 10 Hz) can capture various interesting patterns such as lane changes and sudden jerks while driving straight, which, when contextualized, can provide additional insights to better analyze driving behavior and predict driving risk. Additionally, telematics-based driving risk prediction faces a fundamental challenge in the lack of availability of fine-grained risk labels. Typically, there is no risk data associated with individual trajectories or sub-trajectories. However, a driver's level of riskiness may vary across different driving scenarios, and a model that can capture this variation would be more accurate and useful. Moreover, having risk data available for every trajectory or sub-trajectory would be particularly beneficial for real-time applications, enabling the prevention of dangerous driving actions that could result in accidents.

\bibliographystyle{abbrv}
\bibliography{main}

\end{document}